%% file: vsr_camera_ready.tex
\documentclass[10pt,twocolumn,letterpaper]{article}

\usepackage{iccv}
\usepackage{times}
\usepackage{epsfig}
\usepackage{graphicx}
\usepackage{amsmath}
\usepackage{amssymb}
\usepackage{mathrsfs}
\usepackage{paralist}
\usepackage{lipsum} 
\usepackage[dvipsnames]{xcolor}

\usepackage{multirow}
\usepackage{makecell}
\usepackage{bm}
\usepackage[switch]{lineno}  %

\usepackage{wrapfig}

\usepackage{algorithm}
\usepackage[noend]{algpseudocode}
\usepackage{bbold}
\usepackage{xurl}
\usepackage{booktabs}
\usepackage{xintexpr}% v1.1 or later
\usepackage{authblk}
\usepackage{cuted}

\usepackage[numbers,sort,compress]{natbib}
\usepackage[pagebackref=true,breaklinks=true,colorlinks=true,citecolor=blue,linkcolor=blue,urlcolor=blue,bookmarks=false]{hyperref}

\iccvfinalcopy % *** Uncomment this line for the final submission

\def\iccvPaperID{10703} % *** Enter the ICCV Paper ID here

\ificcvfinal\fi

\colorlet{shadecolor}{gray!40}
 
\begin{document}

%%%%%%%%% TITLE
\title{COMISR: Compression-Informed Video Super-Resolution}
%\title{COMISR: Compression-aware Video super-resolution}

\makeatletter
\renewcommand\AB@affilsepx{, \protect\Affilfont}
\makeatother

% \author[]{Yinxiao Li}
% \author[]{Pengchong Jin}
% \author[]{Feng Yang}
% \author[]{Ce Liu}
% \author[]{Ming-Hsuan Yang}
% \author[]{Peyman Milanfar}
% \affil[]{Google Inc.}
% \affil[]{\textit {\{yinxiao,pengchong,fengyang,celiu,minghsuan,milanfar\}@google.com}}

\author{
Yinxiao Li, Pengchong Jin, Feng Yang, Ce Liu, Ming-Hsuan Yang, Peyman Milanfar \\
\vspace{-4mm}
{\tt\small  {\{yinxiao, pengchong, fengyang, celiu, minghsuan, milanfar\}@google.com}} \\
Google Inc.\\

}

\maketitle
\ificcvfinal\thispagestyle{empty}\fi

%%%%%%%%% ABSTRACT
\begin{abstract}

Most video super-resolution methods focus on restoring high-resolution video frames from low-resolution videos without taking into account compression. 
However, most videos on the web or mobile devices are compressed, and the compression can be severe when the bandwidth is limited. 
In this paper, we propose a new compression-informed video super-resolution model to restore high-resolution content without introducing artifacts caused by compression. 
The proposed model consists of three modules for video super-resolution: bi-directional recurrent warping, detail-preserving flow estimation, and Laplacian enhancement.
All these three modules are used to deal with compression properties such as the location of the intra-frames in the input and smoothness in the output frames.
For thorough performance evaluation, we conducted extensive experiments on standard datasets with a wide range of compression rates, covering many real video use cases.
We showed that our method not only recovers high-resolution content on uncompressed frames from the widely-used benchmark datasets, but also achieves state-of-the-art performance in super-resolving compressed videos based on numerous quantitative metrics.
We also evaluated the proposed method by simulating streaming from YouTube to demonstrate its effectiveness and robustness. 
%
%\footnote{
The source codes and trained models are available at
\href{https://github.com/google-research/google-research/tree/master/comisr}{https://github.com/google-research/google-research/tree/master/comisr}.
%}.

\end{abstract}

\input{1-intro}
\input{2-related_work}
\input{3-method}
\input{4-experiment}

%%%%%%%%% BODY TEXT

\section{Conclusion}

In this work, we propose a compression-informed video super-resolution model which is robust and effective on compressed videos.
Within an efficient recurrent network framework, we design 
three modules to effectively recover more details from the compressed frames.
We conduct extensive experiments on challenging video with a wide range of compression factors.
The proposed COMISR model achieves the state-of-the-art performance on compressed videos qualitatively and quantitatively, while performing well on uncompressed videos.
%
% A key to successfully applying VSR models to se cases is to be robust to the compressed videos, which are widely available from media platforms.
%
%MH: why do you say you set up a framework and evaluation methods for VSR models?
%We strongly believe our work addresses a very practical problem in the VSR research, and sets up a framework and evaluation methods for VSR models in many real applications
%We addresses an important and practical problem for video super-resolution, and sets up a framework and evaluation methods for VSR models in many real applications.

\clearpage
% {\small
% \bibliographystyle{ieee_fullname}
% \bibliography{egbib}
% }

{\small
\bibliographystyle{ieee_fullname}
\bibliography{egbib}
}

\clearpage

%%%%%%%%% TITLE
\twocolumn[  
    \begin{@twocolumnfalse}
        \begin{center}
        \Large\bf{COMISR: Compression-Informed Video Super-Resolution \\ Supplementary Material}
         \end{center}
     \end{@twocolumnfalse}
]
%\title{COMISR: Compression-aware Video super-resolution}

\maketitle
%\ificcvfinal\thispagestyle{empty}\fi

\setcounter{section}{0}
\section{Model Details}

The overall COMISR model is shown in the Figure 2 in the paper manuscript. 
Here we present more details of the two modules: detail-preserving flow estimation and HR frame generate.

\subsection{Detail-Preserving Flow Estimation}
\label{sec:flow_sup}

\begin{figure}[!htb]
\centering
\includegraphics[width=1.0\linewidth]{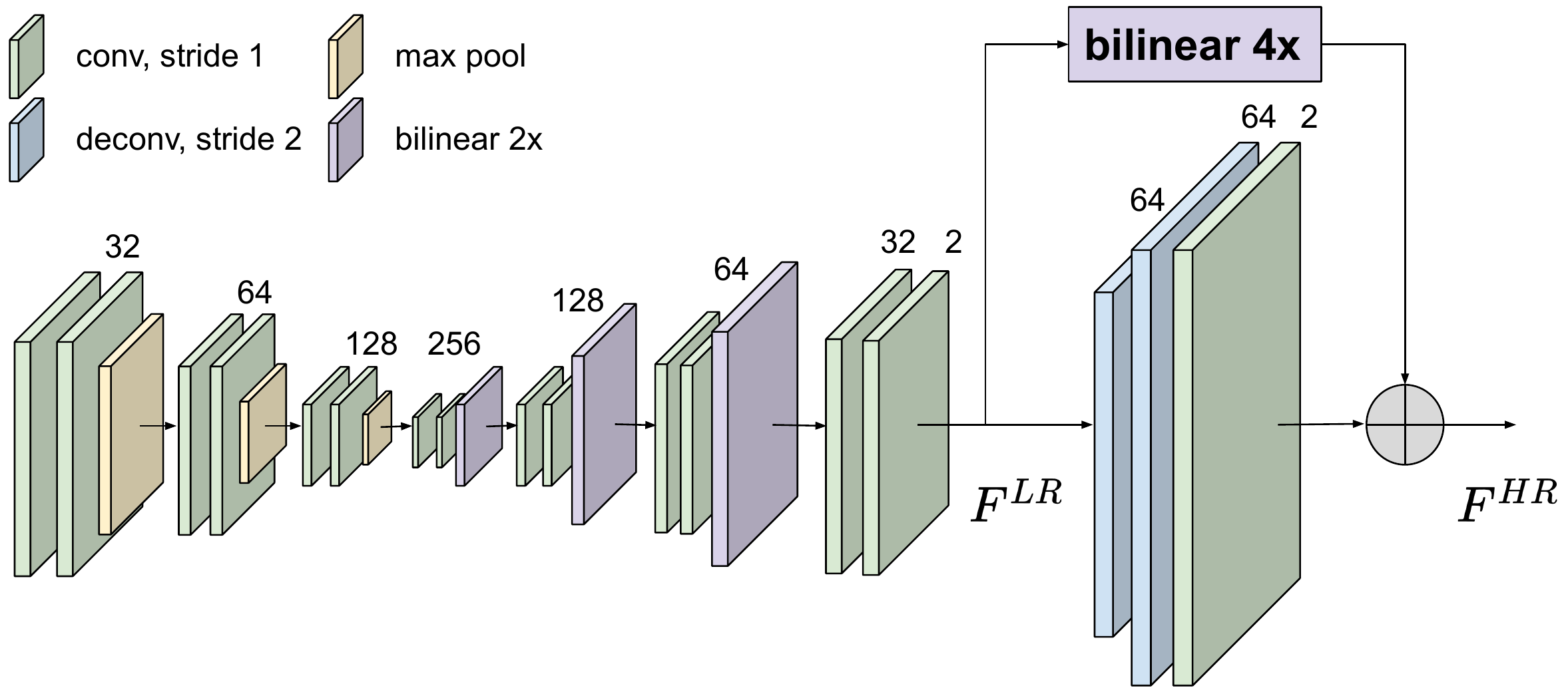}
\caption{Detail-Preserving Flow Estimation}
\label{fig:flow}
\end{figure}

The flow is estimated via a network architecture as is shown in Figure~\ref{fig:flow}.
At each training mini-batch, a short video clip (e.g. 7 frames) is used for training.
The flow estimation can be divided into two parts.
The first part is to learn the motion discontinuities between the consecutive frames.
The second part is for upscaling the estimated flow, which will be then used on the HR frames.
The upscaling process is designed by a learn residual added to a 4$\times$ bilinear upsampling.
Such residual is implemented by repeating a 2$\times$ transpose convolutional layer twice.

All the convolutional kernels are 3$\times$3.
The number of filters in each layer are marked in the Figure~\ref{fig:flow}.
The output flow estimation is used to warp the $t-1$ generated HR frame to the $t$ timestamp, and then used for generating the HR frame in the $t$ timestamp.

\subsection{HR Frame Generator}

As shown in Figure 2 of the manuscript, the input of the HR frame generator is the concatenation of a space-to-depth results of the current estimated HR frame and the current input LR frame.
In the HR frame generator, 10 repeated residual blocks are first employed to extract high-level features.
Then a upscaling module, similar to \emph{Detail-Preserving Flow Estimation} in Section~\ref{sec:flow} is used to create estimated HR frame.

\begin{figure}[!htb]
\centering
\includegraphics[width=0.90\linewidth]{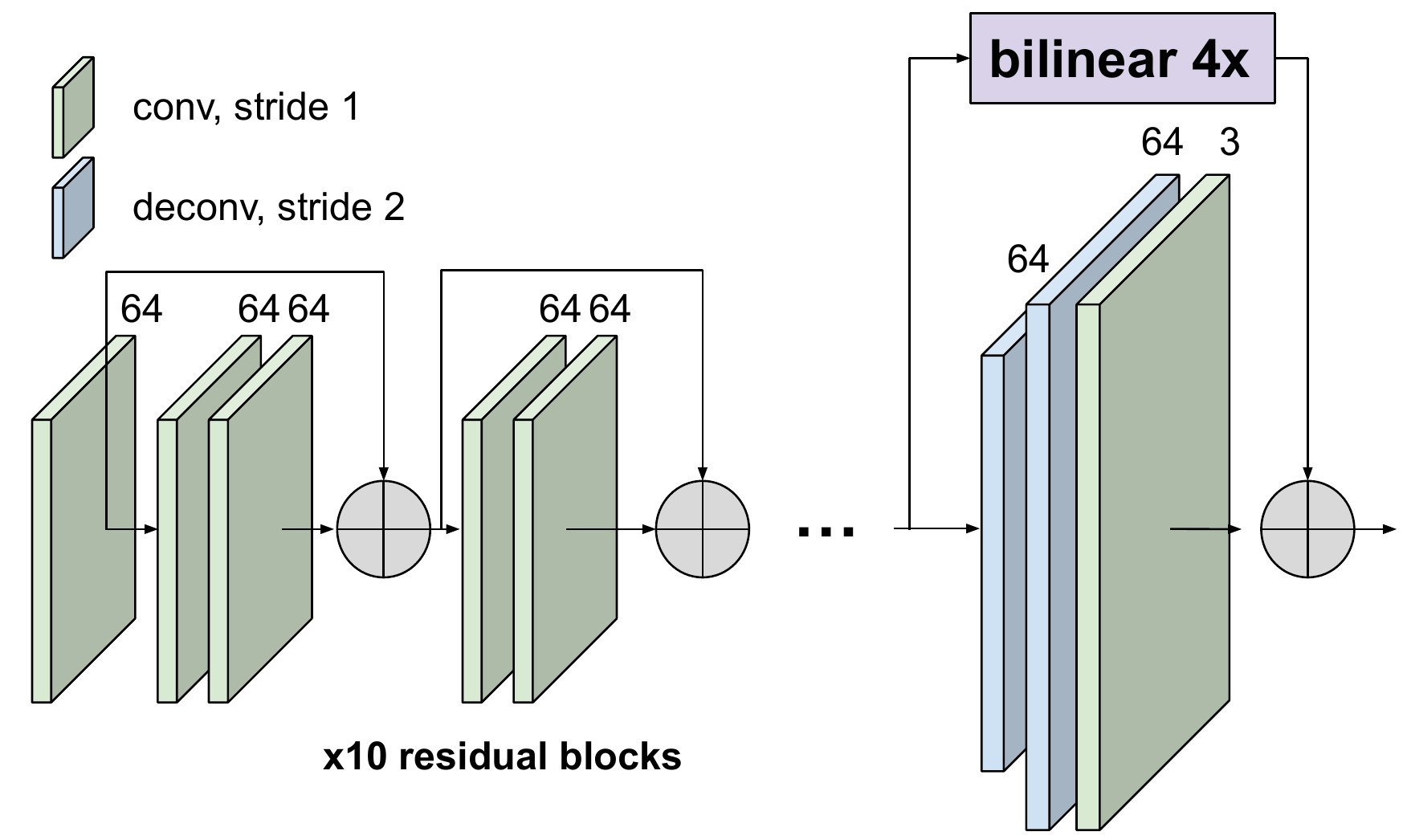}
\caption{HR Frame Generator}
\label{fig:hrnet}
\end{figure}

\section{Experimental Results}

% \subsection{Vimeo-90K-T dataset}
We evaluate the COMISR model against the state-of-the-art VSR methods on the Vimeo-90K-T dataset~\cite{xue2019video}, including FRVSR~\cite{frvsr}, EDVR~\cite{edvr}, TecoGan~\cite{tecogan2020}, and RSDN~\cite{rsdn}.
The Vimeo-90K-T dataset contains 7824 short video clips, where each clip only has 7 frames.
Similar to the observation in ~\cite{rsdn}, the recurrent-based method may not take full advantages due to very short video clips, the COMISR model can still outperform others on the compressed videos.
We show quantitative result below.

\begin{table}[!htb]
\center
\resizebox{0.75\columnwidth}{!}{
\begin{tabular}{c|cc}
     & \hspace{0.5cm}Uncompressed \hspace{0.5cm} & \hspace{0.5cm} CRF25 \hspace{0.5cm} \\  \toprule [0.2em]
 FRVSR~\cite{frvsr}   &  35.64 / 0.932  & 30.07 / 0.788 \\
 TecoGan~\cite{tecogan2020}   &  34.07 / 0.909  & 29.84 / 0.784 \\
 EDVR~\cite{edvr}   &  37.61 / 0.949  & 30.53 / 0.844 \\
  RSDN~\cite{rsdn}   &  37.23 / 0.947  & 29.63 / 0.815 \\
 OURS   &  35.71 / 0.926  & 31.05 / 0.816 \\
\end{tabular}
}
\vspace{1mm}
\caption{Performance evaluation of Y-channel on the Vimeo90K testing set.
}
\label{tbl:vimeo}
\end{table}

% \subsection{Qualitative Comparisons}
% We add additional visual comparison results from the Vid4 and REDS4 datasets.
%
% 

\label{sec:visual_cmp}
\begin{figure*}[!htb]
\centering
\includegraphics[width=1.0\linewidth]{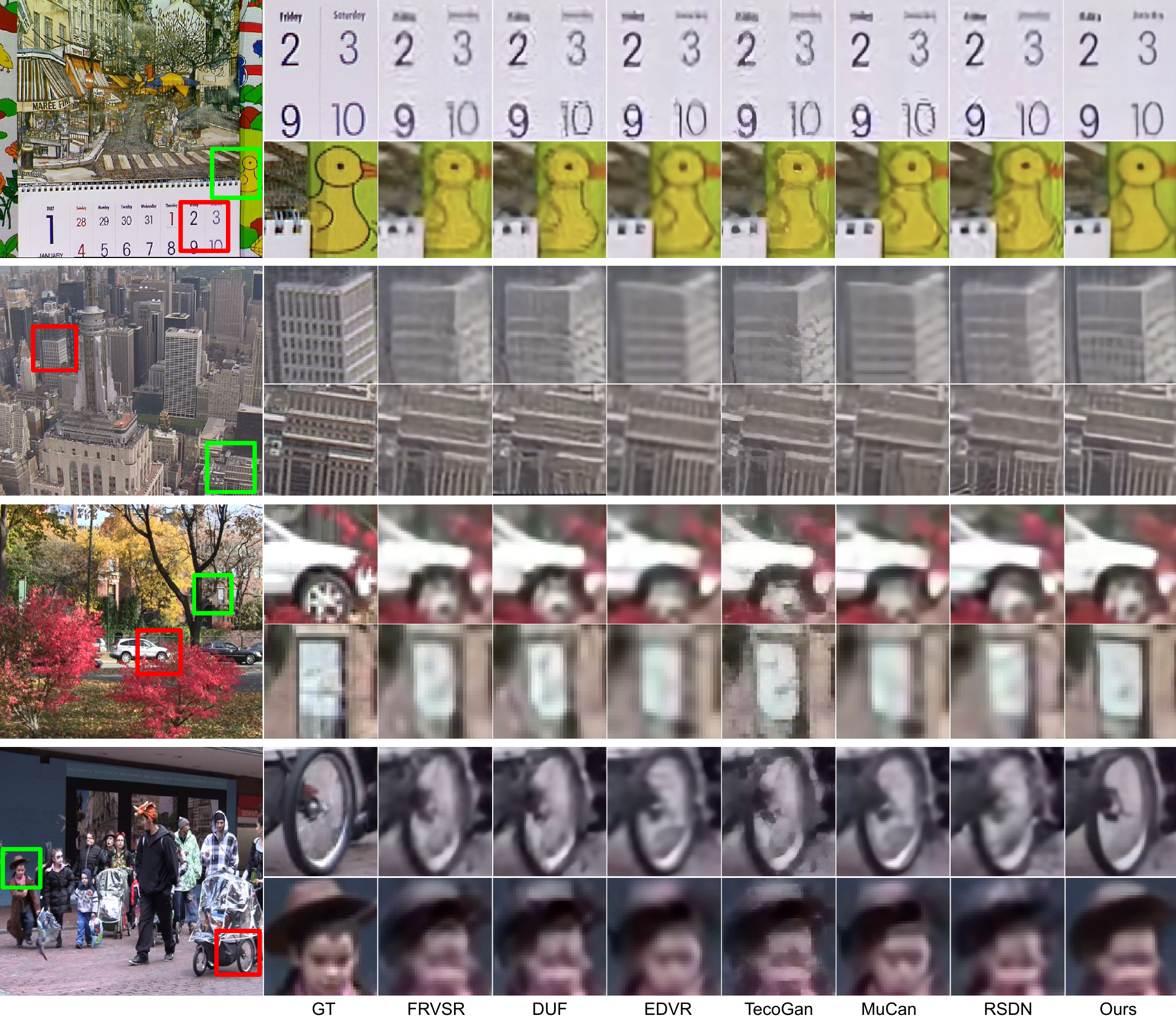}
\caption{Visual example of the Vid4 dataset.
All the LR input frames are compressed with CRF25.
Zoom in for best view.
}
\label{fig:vid4_supp}
\end{figure*}

\begin{figure*}[!htb]
\centering
\includegraphics[width=1.0\linewidth]{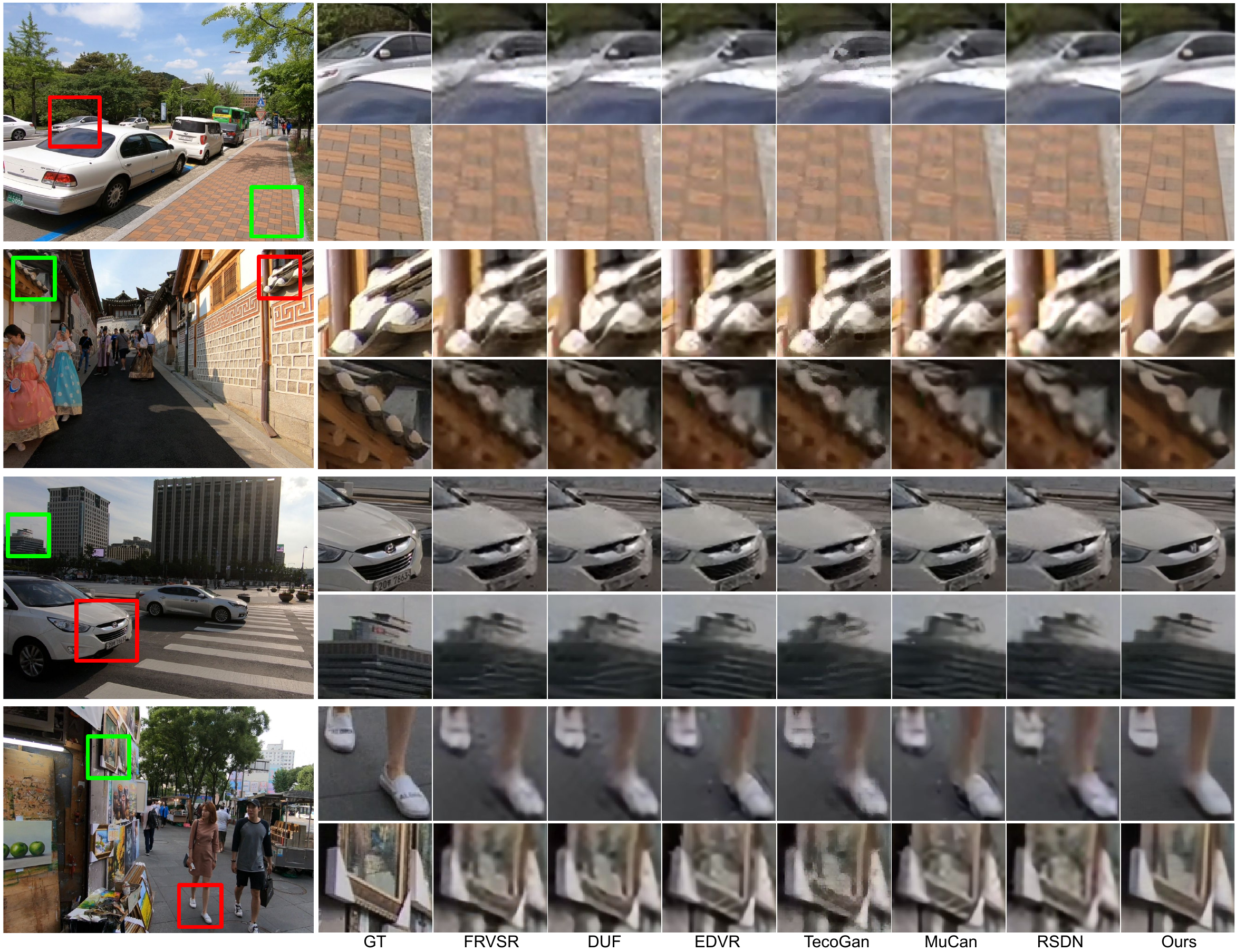}
\caption{Visual example of the REDS4 dataset.
All the LR input frames are compressed with CRF25.
Zoom in for best view.
}
\label{fig:reds4_supp}
\end{figure*}

\end{document}

%% file: 1-intro.tex
\vspace{-4mm}
\section{Introduction}

\begin{figure}[tp]
    \centering
    \includegraphics[width=.98\linewidth]{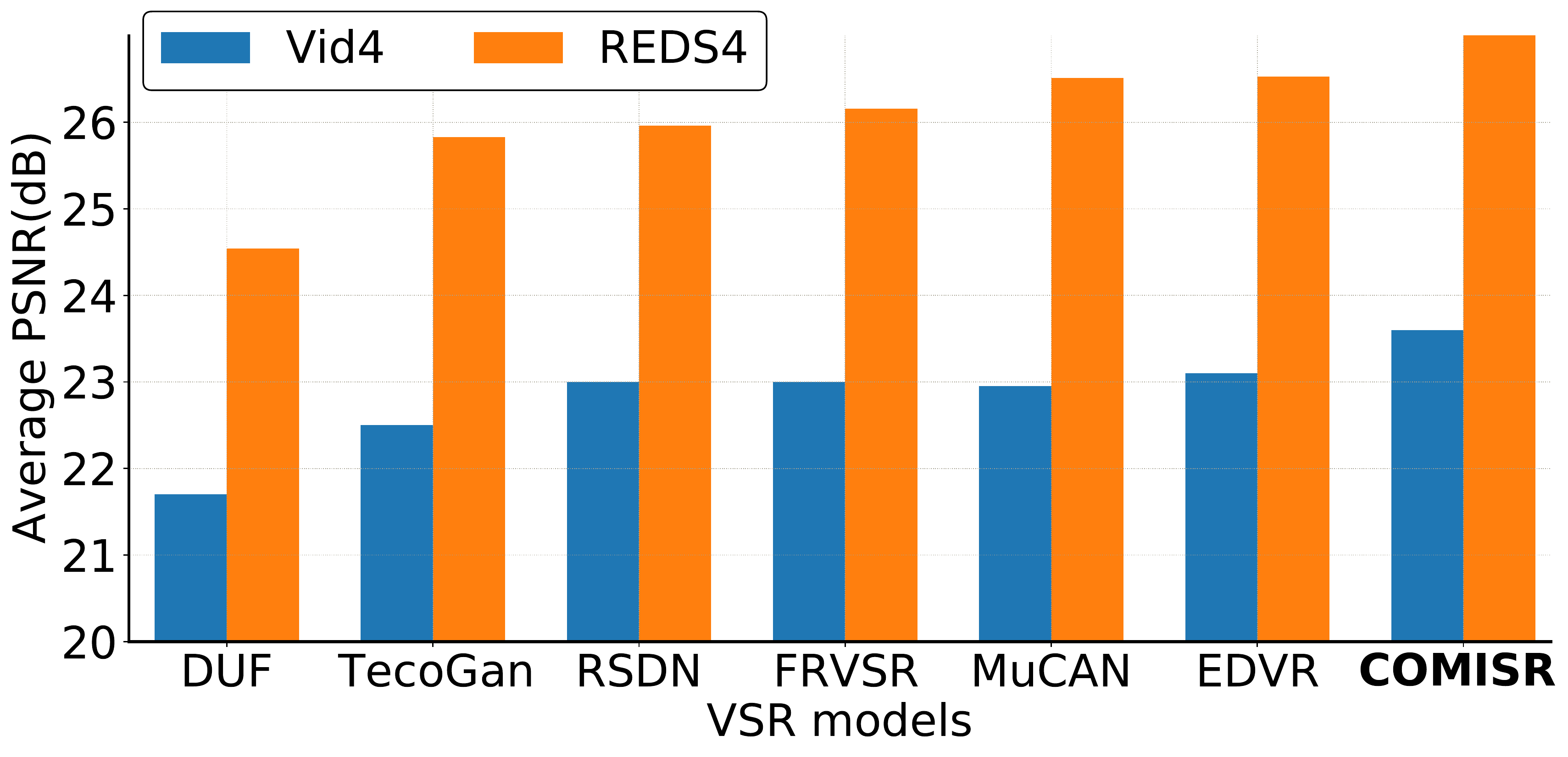}
    \caption{Video super-resolution results ($4 \times$, RGB-channels) on compressed Vid4 and REDS datasets.
    Here we show the results using the most widely adopted compression rate (CRF 23~\cite{ffmpeg}).
    }
    \label{fig:overview_plot}
    \vspace{-2mm}
\end{figure}

%HERE
Super-resolution is a fundamental research problem in computer vision with numerous applications. 
It aims to reconstruct detailed high-resolution (HR) image(s) from low-resolution (LR) input(s).
When the input is one single image, the reconstruction process usually uses learned image priors to recover high-resolution details of the given image, which is called single-image super-resolution (SISR)~\cite{WangTPAMI2020}. 
When numerous frames in a video are available, the reconstruction process uses both image priors and inter-frame information to generate temporally smooth high-resolution results, which is known as video super-resolution (VSR). 
%

%For video super-resolution, the quality of the inputs usually determines the super-resolution results.
%Regarding most of the existing VSR models~\cite{}, the input frames are extracted from the original videos without or with very little amount of compression.  
Although great progress has been made, existing SISR and VSR methods rarely take compressed images as input. 
%
%MH2: bad sentence
% YL: I recommend to keep this sentence. All of our test videos has more or less compression. We need to make it clear that we refer as uncompressed doesn't mean there is no compression but very less compression.
%In this paper, we say that previous work used \textit{uncompressed} videos to emphasize the high-quality, low-compresison-ratio videos.
%
%MH3: see this sentence 
% YL: Thanks for the revision! Looks good to me!
We note that the \textit{uncompressed} videos used in prior work in fact are high-quality image sequences with low compression rate.   
As such, these SR methods tend to generate significant artifacts when operating on heavily compressed images or videos. 
However, most videos on the web or mobile devices are stored and streamed with images compressed at different levels.
For example, a wide-used compression rate (Constant Rate Factor (CRF)) for H.264 encoding is 23 as a trade-off between visual quality and file size. 
%
%MH: too much detail at the beginning
%A common recommended CRF value for visually lossless or near so is 17 or 18. 
%
%A few widely-used multi-media storage platforms even enforce a higher CRF value (between 24-28) for a storage saving.
%
%On the customers side, they would prefer to pay less for the storage for the growing amount of multimedia contents. 
%On the other hand, when the customers retrieve the saved videos, the quality of the video can be restored as much as possible to its original quality using VSR algorithms.  
%
%Given such requirements, we tested a few most recent VSR models on the compressed videos, none of them are able to provide high quality restored outputs upon VSR task. 
%
%For example, the artifacts from the LR frames are amplified in the output HR frames, as shown in Figure~\ref{fig:overview_plot}.
%
%We apply such settings to a few VSR evaluation datasets, and notice that all 
%
%MH2: it is not good to refer to Figure 3 and 4 before Figure 1.
% YL: agree.
%We note the state-of-the-art VSR algorithms do not perform as well when videos are compressed, as shown in Figure~\ref{fig:vid4_sota} and Figure~\ref{fig:reds4_sota}.
We note the state-of-the-art VSR algorithms do not perform well when the input videos are compressed. 

% yinxiao newly added
%One possible solution is applying a denoising model~\cite{Xu_2019_ICCV,Lu_2018_ECCV,Lu_TIP_2019} to remove compression artifacts, followed by one of the state-of-the-art VSR models. 
To handle compressed videos, 
one potential solution is to first denoise images and remove compression artifacts in images~\cite{Xu_2019_ICCV,Lu_2018_ECCV,Lu_TIP_2019} before applying one of the state-of-the-art VSR models. 
At first glance, this is appealing since a VSR model is fed with high-quality frames, similar to directly using the evaluation data, such as Vid4~\cite{vid4}. 
However, our experiments in Section~\ref{denosiervsr} show that this approach would not improve SR results and instead negatively affect the visual quality. 
%
%MH2: it does not "change"...
%With pre-processing, it is likely that the denoising model in the first step will change the degradation kernel used implicitly during the VSR modeling training process. 
% YL: agree with the rephrase.
%
With pre-processing, it is likely that the denoising model in the first step will be significantly different from the degradation kernel used implicitly during the VSR training process.
After the denoising process, the VSR models effectively need to handle more challenging images.  

Another possible solution is to train the existing state-of-the-art VSR models on the compressed images.
%
%MH2:
%This can bring additional compression information to the model training.
% YL: Agree.
This will enforce the VSR models to account for compression artifacts during the training process. 
However, our experiments described in Section~\ref{modelablation} show that simply using compressed frames in model training brings only modest improvement.
In fact, without specific changes to the designs of network modules, such training data may even negatively affect the overall performance.

To address the above-mentioned issues, we propose a compression-informed (i.e., compression-aware)  super-resolution model that can perform well on real-world videos with different levels of compression.
%
%As a first attempt, we start with training on the compressed video data using a few public available models such as FRVSR~\cite{frvsr} and TecoGan~\cite{tecogan2020}, etc.. 
%None of them perform well with such training data. %
%One of the possible explanations is that these previous architecture designs are not robust enough for the compressed inputs. 
%
%Therefore, to design a new architecture that can be robust to the compressed input is critical to such setting. 
Specifically, we design three modules to robustly restore the missing information caused by video compression.
%
%MH: check this sentence YL: If we talk about video compression, we should be able to know which one is i frame. I'd say
% i frame from compressed video frames.
First, a bi-directional recurrent module is developed to reduce the accumulated warping errors from the random locations of the \emph{intra-frame} from compressed video frames~\cite{brnn}. 
Second, a detail-aware flow estimation module is introduced to recover HR flow from compressed LR frames.
Finally, a Laplacian enhancement module is adopted to add high-frequency information to the warped HR frames washed out by video encoding.
We refer to this proposed model as \emph{COMpression-Informed video Super-Resolution (COMISR)}. 

With the proposed COMISR model, we demonstrate the effectiveness of these modules with ablation studies. 
We conduct extensive experiments on several VSR benchmark datasets, including Vid4~\cite{vid4} and REDS4~\cite{Nah_2019_CVPR_Workshops_REDS}, using videos compressed with different CRF values. 
We show that the COMISR model achieves significant performance gain on compressed videos (e.g., CRF23), as shown in Figure~\ref{fig:overview_plot}, and meanwhile maintains competitive performance on uncompressed videos.
In addition, we present evaluation results based on different combinations of a state-of-the-art VSR model and an off-the-shelf video denoiser.
Finally, we validate the robustness of the COMISR model on YouTube videos, which are compressed with proprietary encoders. 

The contributions of this paper can be summarized as: 
\begin{compactitem}
\item We introduce a compression-informed model for super-resolving real-world compressed videos and achieve  state-of-the-art performance. 
\item We incorporate three modules that are novel to VSR to effectively improve critical components for video 
super-resolution on compressed frames.
\item We conduct extensive experiments of state-of-the-art VSR models on compressed benchmark datasets. 
%
%MH: not sure about this sentence. check this one YL: looks good to me.
We also present a new setting for evaluating VSR models on 
YouTube transcoded videos, which is a real-world application scenario that existing evaluation methods do not consider. 
\end{compactitem}

\begin{figure*}[!htb]
\includegraphics[width=.98\linewidth]{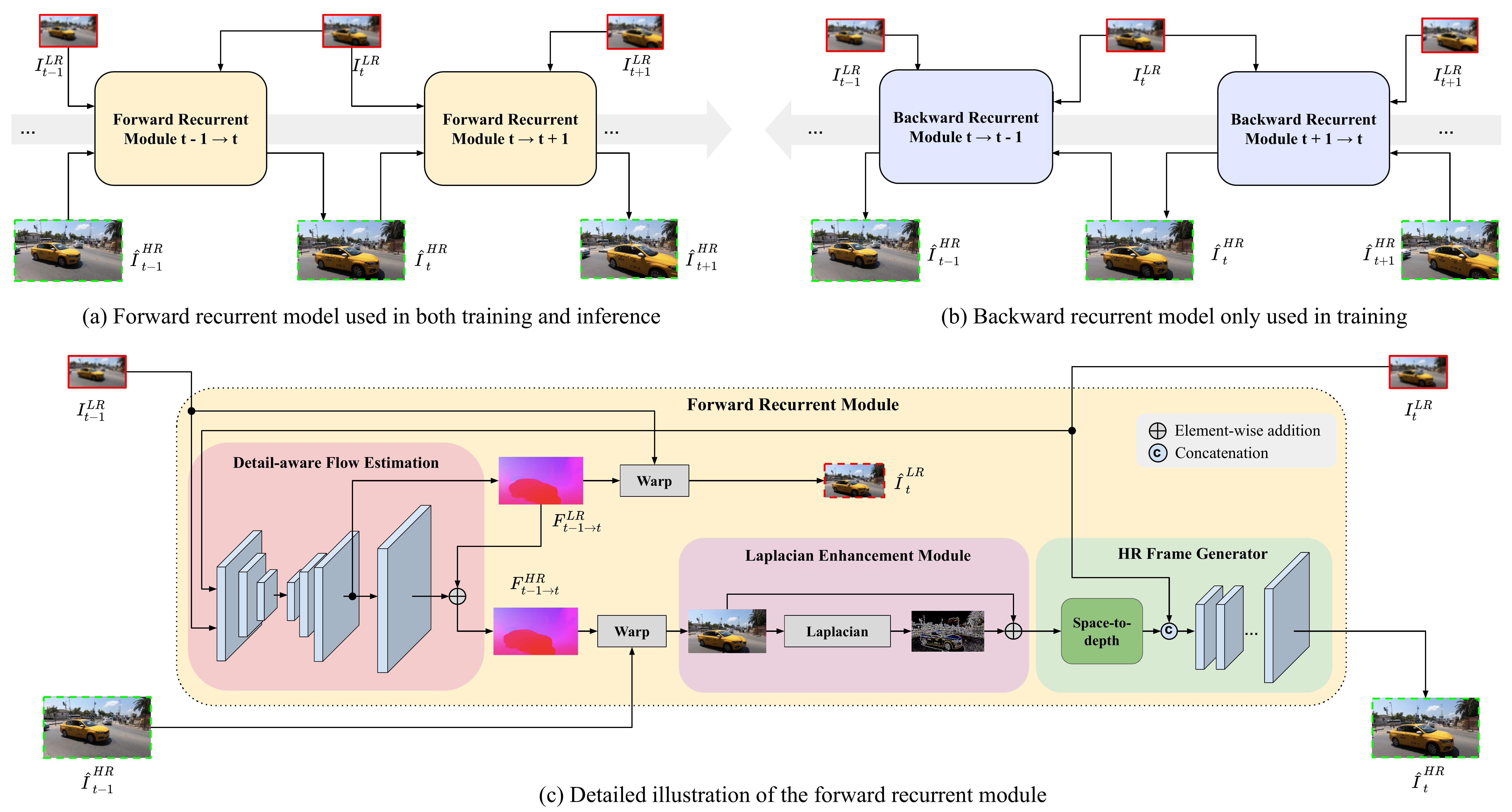}
\caption{Overview of the COMISR model. 
The forward and backward recurrent modules are symmetric and share the weights. 
In the figure, red rectangles represent the LR input frames and green dash-lined rectangles represent the HR predicted frames.}
\label{fig:main_figure}
\vspace{-2mm}
\end{figure*}

% \pengchong{which one is better?}
%\begin{figure*}[!htb]
%\centering
%\includegraphics[width=1.0\linewidth]{figures/cavsr.pdf}
%\caption{Alternative }
%\label{fig:main_figure}
%\vspace{-2mm}
%\end{figure*}

%% file: 2-related_work.tex
%MH: use current tense... no need to use past tense
\section{Related Work}

%WangTPAMI2020,Yang2019TMM,Anwar2020ACMCS
A plethora of super-resolution methods have been developed in the literature based on variational formulations~\cite{SISR-benchmark} or deep neural networks~\cite{WangTPAMI2020,Yang2019TMM,Anwar2020ACMCS}.  
In this section, we discuss recent deep models closely related to our work for super-resolution. 

\subsection{Single-image Super-resolution}
%Single image super-resolution (SISR) aims to recover high-fidelity content from one single input. 

%
Dong~\emph{et al.}~\cite{dong2014eccv} propose the SRCNN model based on convolutional neural networks for single image super-resolution. 
Based on the residual learning framework~\cite{He2015}, Kim~\emph{et al.} propose the VDSR~\cite{kim2016cvpra} and DRCN~\cite{kim2016cvprb} models for more effective image super-resolution. 
To learn more efficient SR models, Dong~\etal~\cite{DONG2016eccv} use a deconvolution layer at the end of the network to directly learn the mapping from low-resolution to high-resolution images.
Similarly, Shi~\etal introduce the ESPCN~\cite{Shi_2016_CVPR} model with an efficient sub-pixel convolution layer at the end of the network. 
In the LatticeNet method~\cite{Luo_ECCV_2020}, a light-weighted model is developed by using a lattice block, which reduces half amount of the parameters while maintaining similar SR performance. 
To learn SR models at multiple scales efficiently, Lai~\emph{et al.}~\cite{Lai_2017_CVPR} develop the LapSRN model which progressively recovers the sub-band residuals of high-resolution images. 
Instead of relying on deeper models, the MemNet~\cite{Tai_2017_ICCV} introduce memory block to exploit long-term dependency for effective SR models.  
On the other hand,   
the SRDenseNet~\cite{Tong_2017_ICCV} and RDN~\cite{Zhang_2018_CVPR} are proposed for SISR based on the DenseNet~\cite{Huang_2017_CVPR} model 
with dense connections.
Haris~\etal~\cite{Haris_2018_CVPR} design a deep back-projection network for super-resolution by exploiting iterative up-sampling and 
down-sampling layers.
In~\cite{Han_2018_CVPR}, the DSRN introduces a dual-state recurrent network model to reduce memory consumption for SISR. 
The MSRN~\cite{Li_2018_ECCV} and RFA~\cite{Liu_2020_CVPR} models use different blocks to efficiently exploit image features.
Recently, attention mechanisms have also been used to improve the super-resolution image quality~\cite{Zhang_2018_ECCV,Dai_2019_CVPR,Mei_2020_CVPR,Niu_ECCV_2020}.

Aside from deep neural network models, generative adversarial networks (GANs) have been adopted for SISR, including
SRGAN~\cite{Ledig_2017_CVPR}, EnhanceNet~\cite{Sajjadi_2017_ICCV}, ESRGAN~\cite{Wang_2018_ECCV_Workshops}, SPSR~\cite{Ma_2020_CVPR} and SRFlow~\cite{Lugmayr_ECCV_2020}.
These methods typically generate visual pleasing results by using adversarial losses~\cite{Goodfellow_2014_NeurIPS} or normalizing flows~\cite{pmlr-v37-rezende15}.
In addition, several models have been developed for SISR based on degrated closer to the real-world scenarios~\cite{Xu_2020_CVPR,Guo_2020_CVPR,Hussein_2020_CVPR,Zhang_2020_CVPR, Wei_ECCV_2020}.
%
%
%Several recent survey papers~\cite{WangTPAMI2020,Yang2019TMM,Anwar2020ACMCS} on SISR are also worth reading.

%\subsection{Video Compression Artifact Removal}

\subsection{Video Super-resolution}
Video super-resolution is a more challenging problem than SISR as both content and motion need to be effectively predicted.
%
%The VSR task is similar to SISR but with another important complication of temporal information between frames.
%
The motion information provides additional cues in restoring high-resolution frames from multiple low-resolution images. 

\vspace{-4mm}
\paragraph{Sliding-window methods.} Multi-frame super-resolution methods potentially can restore more high-resolution details of target frames as more visual information is available. 
On the other hand, these methods need to account for motion content between frames for high quality SR results. 
A number of models compute optical flows between multi-frames to aggregate visual information.
Xue \etal ~\cite{xue2019video}  introduce a task-oriented flow estimation method together with a video processing network for  denoising and super-resolution.
%
%MH: check this sentence... does not sound right. YL: I rephrase this sentence.
Haris \etal ~\cite{RBPN2019}  use multiple back-projected features for iterative refinement rather than explicitly aligning frames. 
Recently, deformable convolution networks~\cite{Dai_ICCV2017_Deformable} have been developed to tackle feature misalignment in dense prediction tasks.
Both EDVR \cite{edvr, wang2020basicsr} and TDAN~\cite{tdan} use deformable convolution models to align features from video frames for video super-resolution. 
Haris~\etal~\cite{STAR2020} design a model that leverages mutually informative relationships between time and space
to increase spatial resolution of video frames and 
interpolate frames to increase the frame rate.
In~\cite{PFNL}, Yi~\etal propose a model that use non-local blocks to fuse spatial-temporal information from multiple frames.
Recenlty, Li \etal~\cite{mucan} present a mutli-correspondence network model to 
exploit spatial and temporal correlation between frames to  
fuse intra-frame as well as iner-frame information 
for video SR. 

\vspace{-4mm}
\paragraph{Recurrent models.}
Recurrent neural networks have been widely used for numerous vision tasks, such as classification~\cite{Liang_2015_CVPR,Donahue2017}, detection~\cite{liu2019looking,Veeriah_2015_ICCV}, and segmentation~\cite{Ventura_2019_CVPR}.
%
%A recurrent model takes the previous outputs as the part of the input while having hidden states. 
%
Such network models can process inputs of any length by sharing model weights across time.
In addition, recurrent models can account for long-range dependence among pixels. 
A number of VSR models have been developed based on recurrent neural networks in recent years.
The FRVSR~\cite{frvsr} model stores the previous information in a HR frame for restoring the current frame in a sequence. 
Fuoli~\cite{fuoli_rlsp_2019} use a
recurrent latent space to encode and propagate temporal information among frames for video super-resolution. 
Most recently, the RSDN model~\cite{rsdn} incorporates a structure-preserving module into a recurrent network and achieves state-of-the-art performance for restoring details from LR frames without relying on motion compensation. 

%% file: 3-method.tex
\section{Proposed Method}
\label{method}

%\subsection{Overview}
%\label{model_overview}

The COMISR model is designed based on a recurrent formulation.
Similar to the state-of-the-art video SR methods~\cite{frvsr,rsdn}, 
it feeds visual information from the previous frames 
to the current one.
The recurrent models usually entail low memory consumption, and can be applied to numerous inference tasks in videos. 

Figure~\ref{fig:main_figure} shows an overview of the COMISR model.
We develop three modules, i.e., bi-directional recurrent warping, detail-aware flow estimation, and Laplacian enhancement modules, to effectively super-resolve compressed videos. 
Given the LR ground truth frames,
we use the forward and backward recurrent modules to generate the HR frame predictions,
and compute content losses against HR ground truth frames in both directions.
In the recurrent module, we predict flows and generate warped frames in both LR and HR,
and train the network end to end using the LR and HR ground truth frames.

\subsection{Bi-directional Recurrent Module}
\label{sec:bidirect}

% \begin{figure}[t]
% \centering
% \includegraphics[width=1.0\linewidth]{figures/birnn.pdf}
% \caption{Bi-directional Recurrent Network}
% \label{fig:bircnn}
% \end{figure}

%NH: different algorithms? one or different algorithms?
One common approach for video compression is to apply different
algorithms to compress and encode frames at different positions in the video stream.
Typically, a codec randomly selects several reference frames, known as the \textit{intra-frames}, and compresses them independently without using information from other frames.
It then compresses the other frames by exploiting consistency and encoding differences from the \textit{intra-frames}.
As a result, the \textit{intra-frames} usually require more bits to encode and have less compression artifacts than the other frames.
Since the locations of \textit{intra-frames} is not known in advance,
to effectively reduce the accumulated errors from the unknown locations of \textit{intra-frames} for video super-resolution,
we propose a bi-directional recurrent network to enforce the forward and backward consistency of the LR warped inputs and HR predicted frames.

Specifically, the bi-directional recurrent network consists of symmetric modules for forward and backward directions.
In the forward direction, we first estimate both the LR flow $F_{t-1 \rightarrow t}^{LR}$ and HR one $F_{t-1 \rightarrow t}^{HR}$ using the LR frames $I_{t-1}^{LR}$ and $I_{t}^{LR}$ (described in Section \ref{sec:flow}).
We then apply different operations separately in LR and HR streams.
In the LR stream, we warp the previous LR frame $I_{t-1}^{LR}$ to time $t$ using $F_{t-1 \rightarrow t}^{LR}$ to obtain the warped LR frame $\tilde{I}_{t}^{LR}$, which will be used at later stages:
\vspace{-0.5mm}
\begin{equation}
\footnotesize
    \tilde{I}_{t}^{LR} = Warp(I_{t-1}^{LR}, F_{t-1 \rightarrow t}^{LR}). 
\vspace{-0.5mm}
\end{equation}
In the HR stream,
we warp the previous predicted frames $\hat{I}_{t-1}^{HR}$ to time $t$ using $F_{t-1 \rightarrow t}^{HR}$ to obtain the warped HR frame $\tilde{I}_{t}^{HR}$,
followed by a Laplacian Enhancement Module to generate accurate HR warped frame:
\vspace{-0.5mm}
\begin{equation}
\footnotesize
    \tilde{I}_{t}^{HR, Warp} = Warp(\hat{I}_{t-1}^{HR}, F_{t-1 \rightarrow t}^{HR}), 
\vspace{-0.5mm}
\end{equation}
\vspace{-0.5mm}
\begin{equation}
\footnotesize
    \tilde{I}_{t}^{HR} = Laplacian(\tilde{I}_{t}^{HR, Warp}) + \tilde{I}_{t}^{HR, Warp}.
\vspace{-0.5mm}
\end{equation}
%MH: what is "space-to-depth" operation? YL: I added explanation in the first time it is mentioned.
We then apply a space-to-depth operation on $\tilde{I}_{t}^{HR}$ to shrink back its resolution while expanding its channel,
fuse it with the LR input $I_{t}^{LR}$ and
pass the concatenated frame to the HR frame generator
to predict the final HR image $\hat{I}_{t}^{HR}$.
We compare $\hat{I}_{t}^{HR}$ with the ground truth HR $I_{t}^{HR}$ to measure the loss. 

Similarly, we apply the symmetric operations in the backward direction to obtain the warped LR frame and the predicted HR frame.
%
%MH: I do not understand "in this case".... YL: as shown in Figure 2, we have backward recurrent module. In such case, the flow and everything will be "reverted". I rephrase the sentence. PTAL.
In this case, the detail-aware flow estimation module generates
the backward flow from time $t$ to $t-1$, and
images are warped by applying the backward flow to the frame at time $t$ for estimating the frame at time $t-1$.

\subsection{Detail-aware Flow Estimation}
\label{sec:flow}

In our recurrent module, we explicitly estimate both the LR and HR flows between neighboring frames and pass this information in forward and backward directions.

Here we take the forward direction for illustration. 
The operations in the backward direction are similarly applied. 
We first concatenate two neighboring LR frames $I_{t-1}^{LR}$ and $I_{t}^{LR}$ and
pass it through the LR flow estimation network to estimate the LR flow $F_{t-1 \rightarrow t}^{LR}$.
Instead of directly upsampling the LR flow $F_{t-1 \rightarrow t}^{LR}$,
we add a few additional deconvolution layers on top of the bilinearly upsampled LR flow.
Thus, a detailed residual map is learned during the end-to-end training, and 
we can better preserve high-frequency details in the predicted HR flow.
%
%MH: check this sentence after moving it to the supplementary material. 
%The detailed operations of the module are shown in Figure \ref{fig:flow}.

% \begin{figure}[t]
% \centering
% \includegraphics[width=1.0\linewidth]{figures/flow.pdf}
% \caption{Detail-Preserving Flow Estimation}
% \label{fig:flow}
% % \vspace{-2mm}
% \end{figure}

%\begin{figure}[t]
%\centering
%\includegraphics[width=0.7\linewidth]{figures/hrnet.pdf}
%\caption{HR Frame Generator}
%\label{fig:hrnet}
%% \vspace{-2mm}
%\end{figure}

\subsection{Laplacian Enhancement Module}

The Laplacian residual has been widely used in numerous vision tasks, including image blending, super-resolution, and restoration.
It is particularly useful at finding fine details from a video frame, where such details could be smoothed out during video compression.
%
%MH2: what do you mean by "retains information and some details"? What is "information? I know you can retain details. 
%YL: information here I refer to texture from the previous frames. I revised as below: detailed textures.
In our recurrent VSR model, the warped predicted HR frame retains detailed texture information learned from the previous frames.
Such details can be easily missing from the up-scaling network, as shown in Figure~\ref{fig:main_figure}.
As such, we add a Laplacian residual to a predicted HR frame to enhance details. 

% \begin{figure}[tp]
% \centering
% \includegraphics[width=0.95\linewidth]{figures/laplacian_cmp.pdf}
% \caption{Laplacian image enhancement. 
% The left image is an intermediate predicted HR frame in the training pipeline.
% The image patch in red box is from the intermediate predicted HR frame.
% The image patch in green box is with the Laplacian image added.}
% \label{fig:laplacian}
%   \vspace{-2mm}
% \end{figure}

An image is enhanced by Laplacian residuals using a Gaussian kernel blur $G(\cdot, \cdot)$ with the width of $\sigma$:
\vspace{-0.5mm}
\begin{equation}
\footnotesize
\tilde{I}_{t}^{HR} = \tilde{I}_{t}^{HR} +  \alpha (\tilde{I}_{t}^{HR} - G(\tilde{I}_{t}^{HR}, \sigma = 1.5)),
\label{eq:laplacian}
\vspace{-0.5mm}
\end{equation}
where $\tilde{I}_{t}^{HR}$ is an intermediate results of the predicted HR frame and $\alpha$ is weighted factor for the residuals.
%
% Figure~\ref{fig:laplacian} shows a comparison of enhancing details using Laplacian images.
% %
% Comparing image patch in red and green boxes, it is clear that the detailed texture has been sharpened.
%
We present more ablation studies in Section~\ref{sec:exp} to demonstrate the effectiveness of Laplacian residuals for enhancing image details. 

By exploiting the Laplacian,
we add details back to the warped HR frame.
%
%MH: what is space-to-depth operation? YL: I added some more explanation here.
This is followed by a space-to-depth operation, which rearranges blocks of spatial data into depth dimension, and then concatenation with the LR input frame.
We pass it through the HR frame generator to obtain the final HR prediction.
%
%MH: remove this sentence when this figure is move to the supplementary material. YL: Done
%The detailed architecture of the HR frame generator is shown in Figure~\ref{fig:hrnet}.

\subsection{Loss Function}
\label{loss_function}
During training, the losses are computed from
two streams for HR and LR frames. 
%
%The losses are designed using both of the streams.
For loss on HR frames, the $\mathcal{L}_{2}$ distance is computed between the final outputs and the HR frames.
In Section~\ref{sec:bidirect}, we describe our bi-directional recurrent module for improving the model quality.
Here, $I_t$ denotes the ground truth frame and $\tilde{I}_t$ denotes the generated frame at time $t$.
For each of the recurrent steps, the predicted HR frames are used to compute losses.
The $\mathcal{L}_{2}$ losses are combined as:  
\vspace{-0.5mm}
\begin{equation}
\footnotesize
    \mathcal{L}_{content}^{HR} = \frac{1}{2N}(\underbrace{\sum_{t=1}^{N}||I_{t}^{HR}-\hat{I}_{t}^{HR}||_2}_\text{forward} + \underbrace{\sum_{t=N}^{1}||I_{t}^{HR}-\hat{I}_{t}^{HR}||_2}_\text{backward}).
\vspace{-0.5mm}
\end{equation}

Each of the warped LR frames from $t-1$ to $t$ is penalized by the $\mathcal{L}_{2}$ distance with respect to the current LR frame,
\vspace{-0.5mm}
\begin{equation}
\footnotesize
    \mathcal{L}_{warp}^{LR} = \frac{1}{2N}(\underbrace{\sum_{t=1}^{N}||I_{t}^{LR}-\tilde{I}_{t-1}^{Warp}||_2}_\text{forward} + \underbrace{\sum_{t=N}^{1}||I_{t}^{LR}-\tilde{I}_{t-1}^{Warp}||_2}_\text{backward}).
\vspace{-0.5mm}
\end{equation}
The total loss is the sum of the HR and LR losses, 
\vspace{-0.5mm}
\begin{equation}
\footnotesize
    \mathcal{L}_{total} = \beta\mathcal{L}_{content}^{HR} + \gamma\mathcal{L}_{warp}^{LR},
\vspace{-0.5mm}
\end{equation}
where $\beta$ and $\gamma$ are weights for each loss.

%% file: 4-experiment.tex
\section{Experiments and Analysis}
\label{sec:exp}
In this section, we first introduce our implementation details and evaluation metrics.
We then evaluate our method against the state-of-the-art VSR models on benchmark datasets.
In addition, we demonstrate that our method performs better 
than a baseline method based on a denoiser and a VSR model.
We also evaluate the COMISR model on real-world compressed YouTube videos.
Finally, we show ablation on the three novel modules with analysis, and user study results.
\subsection{Implementation Details}
\label{impldetails}

\paragraph{Datasets.} 
%Using a public dataset for training VSR models is important so that others can easily reproduce.
%MH: add references for REDS and Viemo (see other papers for references)  YL: done.
We use the REDS~\cite{Nah_2019_CVPR_Workshops_REDS} and Vimeo~\cite{xue2019video} datasets for training. 
The REDS dataset contains more than 200 video sequences for training, each of which has 100 frames with $1280 \times 720$ resolution.
The Vimeo-90K dataset contains about 65k video sequences for training, each of which has 7 frames with $448 \times 256$ resolution.
One main difference between these two datasets is the REDS dataset contains images with much larger motion 
captured from a hand-held device.
To train and evaluate the COMISR model, the frames are first smoothed by a Gaussian kernel with the width of 1.5 and downsampled by a factor of $4$.

%MH: again, cite Vid4 and REDS4 datasets (see other papers) YL: done.
We evaluate the COMISR model on the Vid4~\cite{vid4} and REDS4~\cite{Nah_2019_CVPR_Workshops_REDS} datasets (clip\# 000, 011, 015, 020). 
All the testing sequences contain more than 30 frames.  %which are preferable to evaluate the performance of a recurrent-based model.
%
%MH: the model? do you mean the COMISR model? or all the evaluated models (including other methods)? YL: only our model. I add COMISR
In the following experiments, the COMISR model evaluated on the REDS4 dataset is trained with the REDS dataset using the same setting described in~\cite{edvr}. 
The COMISR model in all the other experiments is trained using the Viemo-90K.

\vspace{-4mm}
\paragraph{Compression methods.} We use the most common setting for the H.264 codec at different compression rates (i.e., different CRF values).  
%
%The value of CRF can be selected between 0 and 51.
The recommended CRF value is between 18 and 28, and the default is 23 (although the CRF value ranges between 0 and 51).
In our experiments, we use CRF of 15, 25, and 35 to evaluate video super-resolution with a wide range of compression rates. 
For fair comparisons, when evaluating other methods, we use the same degradation method to generate the LR sequences before compression.  
Finally, these compressed LR sequences are fed into the VSR models for inference.

\vspace{-4mm}
\paragraph{Training process.} For each video frame, we randomly crop $128\times128$ patches from a mini-batch as input.
Each mini-batch consists of 16 samples. 
The $\alpha$, $\beta$, and $\gamma$ parameters described in Section~\ref{method} are set to 1, 20, 1, respectively.
The model trained with the loss functions described in the Section~\ref{loss_function}. 
We use the Adam optimizer~\cite{adam_opt} with $\beta_{1}=0.9$ and $\beta_{2}=0.999$. 
The learning rate is set to $5\times10^{-5}$.
%
%MH: random compression augmentation? It does not sound right... Check this.  YL: Good catch! I rephrase it a bit. PTAL.
%MH3: this is different from the discussion with you today around 6 pm. I thought you meant it was important to use clean images for VSR although the test images are compressed videos. Check what I revise.
%We adopt video compression as an additional data augmentation method to the training pipeline with a probability of $50\%$ on the input batches.
While we aim to train the COMISR model for VSR with compressed videos as input, we first feed uncompressed images to the model, and 
during the last $20\%$ of the training epochs, 
we randomly add compressed images in the training process with a probability of 50\%.  
%
%MH: randomly selected betwee 15 and 25? Check this.  YL: yes, in the training we randomly picked a CRF value in [15, 25].
The FFmpeg codec is employed for compression with a CRF value randomly selected between 15 and 25.
All the models were trained on 8 NVidia Tesla V100 GPUs.
More details can be found on the project website. 
%MH: I add this sentence to add more credibility. Usually, we can release the code as long as we go through the process. YL: Sounds good to me.
%The source code, datasets, and trained models will be made available to the public. 

\vspace{-4mm}
\paragraph{Evaluation metrics.} 
%MH: unnecessary sentneces 
%The PSNR and SSIM are calculated between the ground truth frames and predicted frames. 
We use PSNR, SSIM, and LPIPS~\cite{zhang2018perceptual} for quantitative evaluation of video super-resolution results. 
%
%The predicted frames are from original frames, compressed frames with CRF 15, 25, and 35, as shown in Section~\ref{compsota}.
%
For the experiments on YouTube videos, we only present video SR results for evaluation since the ground-truth frames are not available.

\subsection{Evaluation against the State-of-the-Arts}
\label{compsota}

\begin{figure*}[!htb]
\centering
\includegraphics[width=1.0\linewidth]{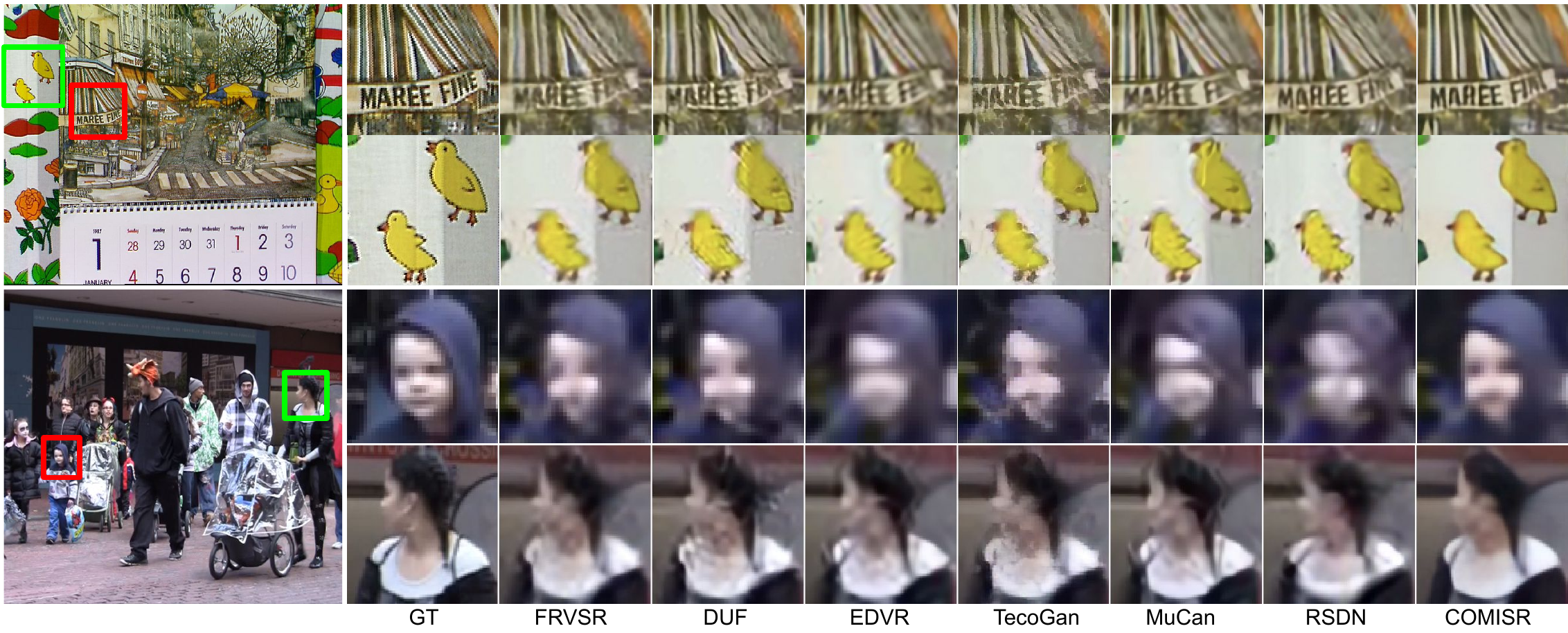}
\caption{Qualitative evaluation on the Vid4 dataset for $4\times$ VSR.
The COMISR model can recover more structure details such as faces and boundaries, with much fewer artifacts. Zoom in for best view.
}
\label{fig:vid4_sota}
\end{figure*}

\begin{figure*}[!htb]
\centering
\includegraphics[width=1.0\linewidth]{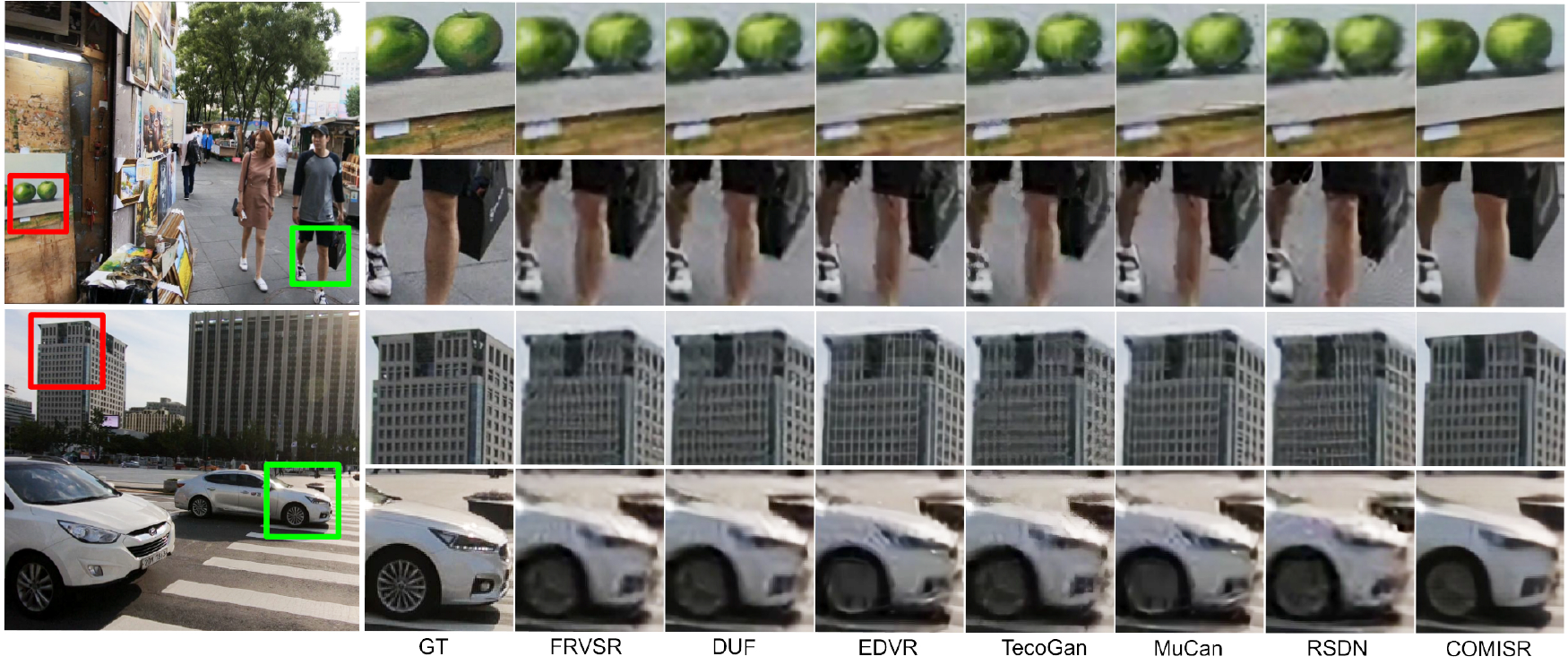}
\caption{Qualitative results on videos from the REDS4 dataset $4 \times$ VSR.
The COMISR model achieves much better quality on detailed textures, with much fewer artifacts. 
%
%MH2: what is 60?  YL: This was from Ce's Photoshop setting. Agree with the current verbal.
%The brightness of the images is increased by 60 for viewing purposes.
The brightness of the images is adjusted for viewing purposes. 
Zoom in for best view.}
\vspace{-0.3cm}
\label{fig:reds4_sota}
\end{figure*}

\begin{table*}[tp]
\centering
\resizebox{2.1\columnwidth}{!}{
\begin{tabular}{l||ccccc||cccc}
& FLOPs & \multicolumn{4}{c||}{CRF 25} &  No compression & \multicolumn{3}{c}{Compressed Results} \\
 Model & \#Param. & calendar  & city  & foliage & walk  & - & CRF15 & CRF25 & CRF35  \\ \toprule [0.2em]
 FRVSR~\cite{frvsr} & \makecell{0.05T\\2.53M} &\makecell{21.55 / {0.631} \\ 19.75 / {0.606}} & \makecell{25.40 / {0.575} \\ 23.79 / {0.572}}  & \makecell{24.11 / {0.625} \\ 24.49 / {0.751}}  & \makecell{26.21 / {0.764} \\ 25.22 / {0.815}}& \makecell{26.71 / {0.820} \\ 25.22 / {0.815}} & \makecell{26.01 / {0.766} \\ 24.38 / {0.753}} & \makecell{24.33 / {0.655} \\ 22.59 / {0.640} }& \makecell{22.05 / {0.482} \\20.35 / {0.469}} \\ \hline
 DUF~\cite{duf}& \makecell{0.62T\\5.82M} & \makecell{21.16 / 0.634 \\ 19.40 / 0.588} & \makecell{23.78 / 0.632 \\ 22.25 / 0.594} & \makecell{22.97 / 0.603 \\ 21.30 / 0.567} & \makecell{24.33 / 0.771 \\ 22.66 / 0.737} & \makecell{27.33 / 0.832 \\ 25.79 / 0.814} & \makecell{24.40 / 0.773 \\ 22.81 / 0.744} & \makecell{23.06 / 0.660 \\ 21.41 / 0.621} & \makecell{21.27 / 0.515 \\ 19.61 / 0.468} \\ \hline
 EDVR~\cite{edvr}& \makecell{0.93T\\20.6M} & \makecell{21.69 / 0.648 \\ 19.87 / 0.599} & \makecell{25.51 / 0.626 \\ 23.90 / 0.586} & \makecell{24.01 / 0.606 \\ 22.27 / 0.570} & \makecell{26.72 / 0.786 \\ 24.89 / 0.754} & \makecell{27.35 / 0.826\\25.85 / 0.808} & \makecell{26.34 / 0.771\\24.67 / 0.740} & \makecell{24.45 / 0.667\\22.73 / 0.627} & \makecell{22.31 / 0.534\\20.62 / 0.487}  \\ \hline
 TecoGan~\cite{tecogan2020}& \makecell{0.14T\\5.05M} & \makecell{21.34 / 0.624\\19.55 / 0.601} & \makecell{25.26 / 0.561\\23.65 / 0.559} & \makecell{23.50 / 0.592\\21.73 / 0.573} & \makecell{25.73 / 0.756\\24.40 / 0.743} & \makecell{25.88 / 0.794\\24.34 / 0.788} & \makecell{25.25 / 0.741\\23.61 / 0.728} & \makecell{23.94 / 0.639\\22.22 / 0.624} & \makecell{21.99 / 0.479\\20.28 / 0.466}     \\ \hline
 MuCAN~\cite{mucan}& - & \makecell{21.60 / 0.643\\19.81 / 0.597} & \makecell{25.38 / 0.620\\23.78 / 0.581} & \makecell{23.93 / 0.599\\22.20 / 0.564} & \makecell{26.43 / 0.782\\24.72 / 0.750} & \makecell{27.26 / 0.822\\25.56 / 0.801} & \makecell{25.85 / 0.753\\24.22 / 0.725} & \makecell{24.34 / 0.661\\22.63 / 0.623} & \makecell{22.26 / 0.531\\20.57 / 0.485}  \\ \hline
 RSDN~\cite{rsdn}& \makecell{0.13T\\6.19M} & \makecell{21.72 / 0.650\\19.89 / 0.599} & \makecell{25.28 / 0.615\\23.68 / 0.575} & \makecell{23.69 / 0.591\\21.94 / 0.554} & \makecell{25.57 / 0.747\\23.91 / 0.711} & \makecell{\colorbox{shadecolor}{\bf{27.92 / 0.851}}\\26.43 / 0.835} & \makecell{\colorbox{shadecolor}{\bf{26.58 / 0.781}}\\24.88 / 0.750} & \makecell{24.06 / 0.650\\22.36 / 0.610} & \makecell{21.29 / 0.483\\19.67 / 0.437}   \\ \toprule [0.2em]
 COMISR & \makecell{0.06T \\ 2.63M} & \makecell{\colorbox{shadecolor}{\bf{22.81 / 0.695}}\\20.39 / 0.667} & \makecell{\colorbox{shadecolor}{\bf{25.94 / 0.640}}\\24.30 / 0.633} & \makecell{\colorbox{shadecolor}{\bf{24.66 / 0.656}}\\22.88 / 0.638} & \makecell{\colorbox{shadecolor}{\bf{26.95 / 0.799}}\\25.21 / 0.788} & \makecell{27.31 / 0.840\\25.79 / 0.835} & \makecell{26.43 / 0.791\\24.76 / 0.778} & \makecell{\colorbox{shadecolor}{\bf{24.97 / 0.701}}\\23.21 / 0.686} & \makecell{\colorbox{shadecolor}{\bf{22.35 / 0.509}}\\20.66 / 0.494}   \\ \hline
\end{tabular}
}
\vspace{1mm}
\caption{Performance evaluation on compressed Vid4 videos. 
For each entry, the first row is PSNR/SSIM on Y channel, and the second row is PSNR/SSIM on RGB channels. 
The best method on the Y channel for each column is highlighted in bold and shade. 
The FLOPs are reported based on the Vid4 $4 \times$ VSR. The FLOPs and \#Param of FRVSR is based on our implementation.}
\label{tbl:vid4_compare}
\vspace{-2mm}
\end{table*}

\begin{table*}[tp]
\centering
\resizebox{2.1\columnwidth}{!}{
\begin{tabular}{l||ccccc||cccc}
&  & \multicolumn{4}{c||}{CRF 25} &  No compression & \multicolumn{3}{c}{Compressed Results} \\ 
 Model & \#Frame & clip\_000  & clip\_011  & clip\_015 & clip\_020  & - & CRF15 & CRF25 & CRF35  \\ \toprule [0.2em]
 FRVSR~\cite{frvsr} & recur(2) & 24.25 / 0.631 & 25.65 / 0.687  & 28.17 / 0.770 & 24.79 / 0.694 & 28.55 / 0.838 & 27.61 / 0.784 & 25.72 / 0.696& 23.22 / 0.579 \\ 
 DUF~\cite{duf}& 7 & 23.46 / 0.622 & 24.02 / 0.686 & 25.76 / 0.773 & 23.54 / 0.689 & 28.63 / 0.825 & 25.61 / 0.775 & 24.19 / 0.692 & 22.17 / 0.588 \\
 EDVR~\cite{edvr}& 7 & 24.38 / 0.629 & 26.01 / 0.702 & 28.30 / 0.783 & 25.21 / 0.708 & \colorbox{shadecolor}{\bf{31.08 / 0.880}} & \colorbox{shadecolor}{\bf{28.72 / 0.805}} & 25.98 / 0.706 & 23.36 / 0.600  \\ 
 TecoGan~\cite{tecogan2020}& recur(2) & 24.01 / 0.624 & 25.39 / 0.682 & 27.95 / 0.768 & 24.48 / 0.686 & 27.63 / 0.815 & 26.93 / 0.768 & 25.46 / 0.690 & 22.95 / 0.589     \\ 
 MuCAN~\cite{mucan}& 5 & 24.39 / 0.628 & 26.02 / 0.702 & 28.25 / 0.781 & 25.17 / 0.707 & 30.88 / 0.875 & 28.67 / 0.804 & 25.96 / 0.705 & 23.55 / 0.600  \\
 RSDN~\cite{rsdn}& recur(2) & 24.04 / 0.602 & 25.40 / 0.673 & 27.93 / 0.766 & 24.54 / 0.676 & 29.11 / 0.837 & 27.66 / 0.768 & 25.48 / 0.679 & 23.03 / 0.579   \\ \toprule [0.2em]
 
 COMISR & recur(2) & \colorbox{shadecolor}{\bf{24.76 / 0.660}} & \colorbox{shadecolor}{\bf{26.54 / 0.722}} & \colorbox{shadecolor}{\bf{29.14 / 0.805}} & \colorbox{shadecolor}{\bf{25.44 / 0.724}} & 29.68 / 0.868 & 28.40 / 0.809 & \colorbox{shadecolor}{\bf{26.47 / 0.728}} & \colorbox{shadecolor}{\bf{23.56 / 0.599}}  \\ \hline
\end{tabular}
}
\vspace{1mm}
\caption{Performance evaluation on compressed the REDS4 dataset. 
Each entry shows the PSNR/SSIM on RGB channels.
The best method for each column is highlighted in bold and shade, and recur(2) indicates a recurrent network using 2 frames.}
\label{tbl:reds4_compare}
\vspace{-2mm}
\end{table*}

We evaluate the COMISR model against state-of-the-art VSR methods, including FRVSR~\cite{frvsr}, DUF~\cite{duf}, EDVR~\cite{edvr}, TecoGan~\cite{tecogan2020}, MuCAN~\cite{mucan}, and RSDN~\cite{rsdn}.
Three of the evaluated methods are based on recurrent models, whereas the other three use temporal sliding windows (between 5 and 7 frames).
%
%All the implementations and trained models are from the public available code; if not, implemented by us with our best understanding of the paper.
%MH: you can be clear about the ones you use the original code and those you implement. 
When available, we use the original code and trained models, and otherwise implement these methods.
%
%For fair comparisons, we use the same degradation methods, when available.  
%
For fair comparisons, the LR frames have been generated the same as described in the published work. 
These LR frames are then compressed and fed into the super-resolution networks for performance evaluation.

For the Vid4 dataset~\cite{vid4}, the PSNR and SSIM metrics are measured on both the Y-channel and RGB-channels, as shown in Table~\ref{tbl:vid4_compare}. 
We present the averaged performance on uncompressed videos (original sequences), and videos compressed at different levels (CRF15, 25, 35).
%
%MH: what do you want to say here? Remove it? YL: I want to emphasis on e.g. column 2-5 in Table 2.
%MH2: I do not know why you have this sentence since you mentioned you have results on CRF15, 25, and 35 in the sentence above. 
% YL: Because we want to emphasis that the individual sequence is also better.
We also report the individual sequence performance under CRF25.
%MH: I add this line. YL: SG. I plan to add more visual results.
More results on other CRF factors are presented in the supplementary material. 
Overall, the COMISR method outperforms all the other methods on videos with medium to high compression rates by 0.5-1.0db in terms of PSNR. 
Meanwhile, our method performs well (2nd or 3rd place) in less compressed videos.
Figure~\ref{fig:vid4_sota} shows some results by the evaluated methods from two sequences. 
The COMISR model can recover more details from the LR frames with fewer compression artifacts.
Both quantitative and visual results show that the COMISR method achieves the state-of-the-art results on compressed videos. 

We also evaluate the COMISR model against the state-of-the-art methods on the REDS4 dataset~\cite{Nah_2019_CVPR_Workshops_REDS}.
Unlike the Vid4 dataset, the sequences in this set are longer (100 frames) and more challenging with larger movements between frames. 
%
%MH2: too many typos..
%Table~\ref{tbl:reds4_compare} show the COMISR model achieves the best performance on the compressed videos from the REDS4 dataset.
%
Table~\ref{tbl:reds4_compare} shows the COMISR model achieves the best performance on the compressed videos from the REDS4 dataset.
%, similar to the Vid4 dataset.
%
Figure~\ref{fig:reds4_sota} shows that 
our method is able to recover more details such as textures from the bricks on the sidewalk and windows on the buildings.

It is known that low-level structure accuracy (e.g., PSNR or SSIM) does not necessarily correlate well with high-level 
perceptual quality.
In other words, perceptual distortion cannot be well characterized by such low-level structure accuracy~\cite{blau2018}.
%
%Recent metrics based on deep learning feature maps such as LPIPS~\cite{zhang2018perceptual} can capture more meaningful perceptual similarity measure between frames. 
%
We also use the LPIPS~\cite{zhang2018perceptual} for performance evaluation. 
%
%Such evaluation metric is more important in restoration of the compressed videos, simply because the output HR frames always contain erroneous perceptual contents, as shown in Figure~\ref{fig:vid4_sota} and Figure~\ref{fig:reds4_sota}. 
%
%Therefore, we also evaluate the perceptual quality of the output HR frames from the compressed inputs using LPIPS metric, as shown in 
Table~\ref{tbl:lpips} shows the evaluation results using the LPIPS metric on both Vid4 and REDS4 datasets. 
Overall, the COMISR model performs well against the state-of-the-art methods on both datasets using the LPIPS metric.
%(either second best on the Vid4 dataset or best on the more challenging REDS4 dataset).
%

\begin{table}[tp]
\resizebox{1.0\columnwidth}{!}{
\begin{tabular}{c|ccccccc}
     & FRVSR  & TecoGan  & DUF & EDVR & MuCAN & RSDN & COMISR  \\ \toprule [0.2em]
 Vid4& 4.105&\colorbox{shadecolor}{\bf{3.245}}&4.010&4.396&3.985&4.292&3.689  \\
 REDS4 & 4.188&3.643&4.223&4.075&4.085&4.423&\colorbox{shadecolor}{\bf{3.384}} \\
\end{tabular}
}
\vspace{1mm}
\caption{Performance evaluation using the LPIPS~\cite{zhang2018perceptual} metric (lower is better). Our method performs well, especially on the more challenging REDS4 dataset. 
}
\label{tbl:lpips}
\vspace{-2mm}
\end{table}

We show video super-resolution results on the project website. 
Although the compression artifacts are not easily observable in the LR frames, 
such artifacts are amplified and easily observed after super-resolution.
For the compressed videos, the COMISR model
effectively recovers more details from the input videos with fewer artifacts.
%than the state-of-the-art methods. 
%
%MH2: add a CRF value in CRFXX
%YL: Add CRF35. Even we are SOTA on CRF35, I think we can still say "doesn't perform well on..."?
The COMISR model does not perform well on highly compressed (e.g., CRF35) videos. 
Some failure cases are due to heavy compression so that necessary details are missing for super-resolving frames.
Other failure cases are caused by extremely large movements 
in the videos.

\begin{figure*}[!htb]
\centering
\includegraphics[width=1.0\linewidth]{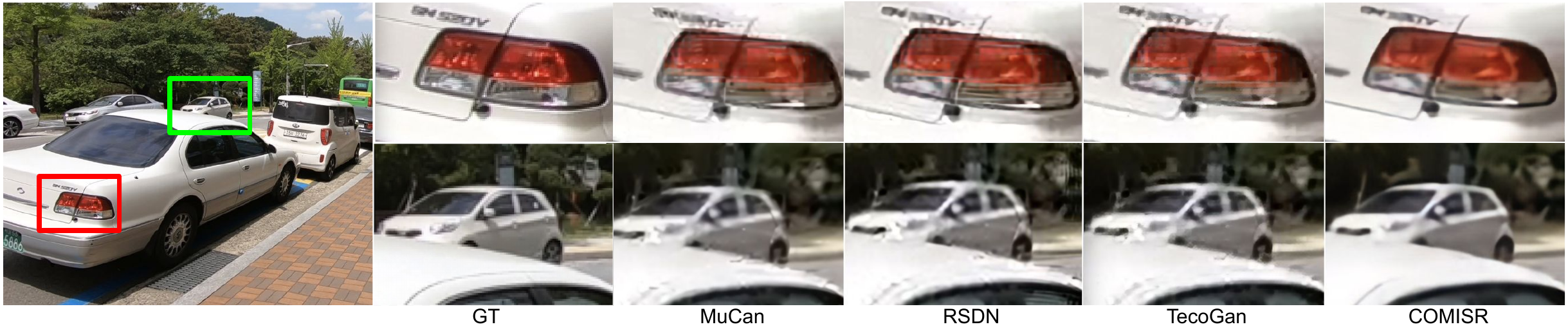}
\caption{$4 \times$ VSR results on REDS4 videos downloaded from YouTube with resolution of 360 pixels. Zoom in for best view.}
\label{fig:reds4_360p}
\vspace{-0.3cm}
\end{figure*}

\subsection{VSR on Denoised Videos}
\label{denosiervsr}
As shown in Figure~\ref{fig:vid4_sota} and Figure~\ref{fig:reds4_sota}, the COMISR model generates high-quality frames with fewer artifacts from compressed videos. 
An interesting question is whether the state-of-the-art methods can achieve better results if the compressed videos are first denoised. 
As such, we use the state-of-the-art compressed video quality method, STDF~\cite{stdf}, for evaluation.

Using the settings described in Section~\ref{compsota}, we compress video frames with CRF25.
The STDF method is then used to remove the compression artifacts and generate enhanced LR frames as inputs for the state-of-the-art VSR methods.  
%
%Such LR frames will be used for evaluating the state-of-the-art methods.
%
%MH: qualitative? YL: thanks for catching!
%We report the qualitative results of VSR only and video denoiser + VSR on both Y and RGB channels in Table~\ref{tbl:denoiser_vsr}
Table~\ref{tbl:denoiser_vsr} shows
the quantitative results by the COMISR model and the state-of-the-art VSR methods on videos denosied by the STDF scheme.
We note that the performance of all of the evaluated method drops on the denoised LR frames.
%
%MH: check this sentence
This can be attributed to that a separate denoising step is not compatible with the learned degradation kernel from the VSR methods.
%
%MH: check this sentence
%MH2: I still do not quite understand this sentence. What is "compression augmentation"?  I do not understand Section 4.5
%In addition, as reported in  Section~\ref{modelablation}, simply adding compression augmentation to the existing model training would not perform well.
%YL: "compression augmentation" is adding compression or not adding compression (randomly) to the input training sequence. The CRF is also random in [15-35] etc.
%
%MH3: again, this is different from what I understand from discussion with you today. See the revised sentence.
%In addition, as discussed in Section~\ref{modelablation}, simply adding compression augmentation to the existing model training would not perform well.
%YL: Looks good to me. Thanks!
In addition, as discussed in Section~\ref{modelablation}, simply using compressed images for model training does not lead to good VSR performance. 
%
%MH: I do not understand "standalone" model... Revise it? YL: I want to emphasis on our one model can cover both uncompressed and compressed videos. I rephrase the sentence. PTAL.
These results show that the COMISR model is able to efficiently recover more details from compressed videos, and outperforms state-of-the-art models on denoised videos.

\begin{table}[tp]
\resizebox{1.0\columnwidth}{!}{
\begin{tabular}{c|cccc}
            & \multicolumn{2}{c}{VSR only} & \multicolumn{2}{c}{Video Denoiser + VSR} \\
 Model &     Y-Channel      &   RGB-Channels       &   Y-Channel        &   RGB-Channels       \\ \toprule [0.2em]
 EDVR  &    24.45 / 0.667   &  22.73 / 0.627        &  22.56 / 0.581         &   20.94 / 0.541       \\
 TecoGan &  23.94 / 0.639 & 22.22 / 0.624 & 22.25 / 0.541 & 20.63 / 0.530         \\
 MuCan &   24.34 / 0.661 & 22.63 / 0.623 & 22.47 / 0.577 & 20.87 / 0.538        \\
 RSDN &  24.06 / 0.650 & 22.36 / 0.610 & 22.19 / 0.560 & 20.59 / 0.520  \\ \hline  
 COMISR &  \colorbox{shadecolor}{\bf{24.97 / 0.701}} & \colorbox{shadecolor}{\bf{23.21 / 0.686}} &  - & -  \\ \hline  
\end{tabular}
}
\vspace{1mm}

\caption{Ablation study on applying a video denoiser to the compressed frames before the VSR models using the Vid4 dataset.
%MH: the sentnece is not correct. You have both Y and RGB channels. Check this. YL: thanks for catching! 
%Each entry shows the PSNR/SSIM on Y-channel.
%
Each entry shows the PSNR/SSIM results on the Y or RGB channel.
The COMISR model outperforms the state-of-the-art VSR methods with the STDF~\cite{stdf} denoiser.
}
\vspace{-2mm}
\label{tbl:denoiser_vsr}
\end{table}

\subsection{Evaluation on Real-World Compressed Videos}
\label{youtuberesults}
%Compressed videos are widely available. 
Most videos on the web are compressed where
frames can be preprocessed by a combination of proprietary   methods.
%
%Therefore, we design a new experiment to demonstrate the robustness of our proposed method.
%
We use the videos from the REDS4 testing dataset for experiments as the image resolution is higher.

We first generate uncompressed videos out of the raw frames, and then upload them to YouTube.
These videos are encoded and compressed at different resolutions for downloading.
In our setting, the uploaded videos are of $1280\times720$ pixels.
%
%MH2: images have two dimensions. 480 \times x?, 360 \times X?
% YL: This is youtube standard. See here: https://screenshot.googleplex.com/9KCD8DqANNaajLS. 
%MH3: I still do not understand. Unless an image is a sqaure, otherwise, this 480, 360, etc, only refers to one dimension. Or it means transmission rate.
%MH3: I see... it means bit rate.. https://support.google.com/youtube/answer/2853702?hl=en#zippy=%2Cp-fps%2Cp It is best to add a reference here. 
%
The resolutions that are available for downloading on YouTube are 480p, 360p, 240p, and 144p.
% The bitrates that are available for downloading on YouTube are 480p, 360p, 240p, and 144p.
%
%MH2: 360 pixels? you need to say 360 \times x?
% YL: please see above.
In the following experiments, we download the videos at 360p using the YouTube-dl~\cite{youtube-dl}.
We evaluate three state-of-the-art methods, including MuCAN~\cite{mucan}, RSDN~\cite{rsdn}, and TecoGan~\cite{tecogan2020} on these videos that are compressed by proprietary methods by YouTube. 
Figure~\ref{fig:reds4_360p} shows the VSR results by the evaluated methods, where the COMISR model produces better visual results with less artifacts.
%MH2: can add one sentence for conclusion on Figure 5, or expand the last sentence. 
% YL: I expand the sentence.
%
%Since there is no ground truth on the output HR frames, we here only com-pare the visual quality, as is shown in Figure~\ref{fig:reds4_360p}. 
%MH: it does not sound right... as you just mention that there is no ground truth...
% We will conduct a user study on more videos and report the visual and quantitative results in the supplementary materials. check this sentnece. 
% More videos and user study results are presented in the project website.

\subsection{Ablation Study}
\label{modelablation}
We analyze the contribution of each module in the COMISR model.
We start with the recurrent module described in Section~\ref{method} as the baseline model. 
Similar to FRVSR~\cite{frvsr}, the recurrent model computes the flow between consecutive frames, warps the previous frame to the current, and upscales the frames. 
%
%MH2: in Section 4.1, there is no such word "compression augmentation". In fact, I have not seen this term before. Can you tell me who uses similar terms (it not the same one)?
% YL:  "compression augmentation" is adding compression or not adding compression (randomly) to the input training sequence. The CRF is also random in [15-35] etc. Shall we say "compression as an augmentation" here?
%MH3: I think you mean here we use uncompressed and compressed images for training. However, how do you use them for training? I will talk to you on this. 
We carry out two sets of ablation studies, with or without using compressed images, to show the effectiveness of each module (see Section~\ref{impldetails}).

%MH2: I need to read this section once I understand what you mean by "compression augmentation". 
Table~\ref{tbl:model_ablation} shows the ablation studies where we incrementally add each module to the basic recurrent model.
For each setting, the model is trained with and without compressed images, and then evaluated on original and compressed frames.
The results show that each module helps achieve additional performance gain, in both training process with only compressed images or a combination of compressed and uncompressed images. 
We note it is important to add some uncompressed images in the training process to achieve best results on compressed videos. 
%
%With the three novel modules, the gain becomes much larger.
The full COMISR model performs best among all settings. 
%
%MH: I do not understand this sentence. YL: I was tring to say adding compression in trainig will make the eval number on uncompressed video drop a bit but not too much. I rephrase the below sentence.
%MH3: it is not good to say aug CRF15-25 or No compression Aug in Table 5 entries. Instead, you can say "Compressed Images Only" and "Compressed and Uncompressed images" (can use smaller font). I am not sure why you say "uncompressed" and "CRF25" in the second row ...
%Adding uncompressed images slightly affects the performance of the COMISR model on the original input videos. 
%
For example, the fourth row in Table~\ref{tbl:model_ablation}, the uncompressed PSNR on Vid4 drops 0.17 dB.

\begin{table}[tp]
\resizebox{1.0\columnwidth}{!}{
\begin{tabular}{c|cccc}
            & \multicolumn{2}{c}{No compression Aug} & \multicolumn{2}{c}{Aug CRF15-25} \\
 Components &   Uncompressed      &   CRF25       &   Uncompressed    &   CRF25       \\ \toprule [0.2em]
 Recur  &    26.61 / 0.808   &  23.97 / 0.634         &  26.53 / 0.815         &   24.23 / 0.648       \\
 Recur + a &   27.16 / 0.837 &  24.24 / 0.650       &    26.64 / 0.818   &  24.74 / 0.686          \\
 Recur + ab&   27.45 / 0.844  &  24.27 / 0.649    &   27.27 / 0.838        &  24.92 / 0.696        \\
 Recur + abc&  27.48 / 0.845  & 24.31 / 0.650     &  27.31 / 0.840 & 24.97 / 0.701  \\ \hline      
\end{tabular}
}
\vspace{1mm}
\caption{Ablations on three modules of the COMISR model on Vid4: (a) bi-directional recurrent module, (b) detail-aware flow estimation, and (c) Laplacian enhancement module.
Each entry shows the PSNR/SSIM values on the Y-channel.
}
\label{tbl:model_ablation}
\vspace{-0.1cm}
\end{table}

\subsection{User Study}
\label{userstudy}
\begin{figure}[!htb]
\centering
\includegraphics[width=1.0\linewidth]{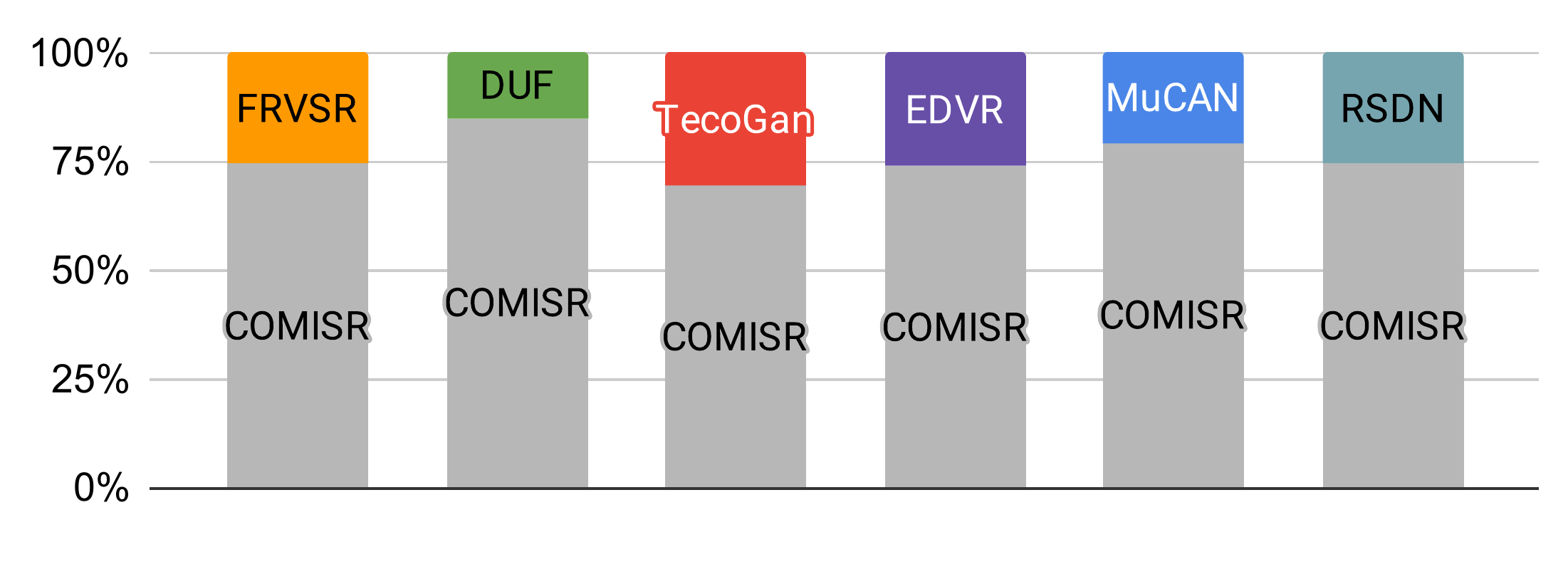}
\caption{Aggregated user study results on Vid4 and Reds4.
Results show that users favored COMISR against all other compared methods.
}
\vspace{-0.2cm}
\label{fig:userstudy}
\end{figure}

%MH2: there are so many typos... You need to pay attention to details...
%To better evaluate the visual quality of the generate HR videos, we conduct a user study using Amazon MTurk~\cite{mturk} on the Vid4~\cite{vid4} and REDS4~\cite{Nah_2019_CVPR_Workshops_REDS}.
To better evaluate the visual quality of the generated HR videos, we conduct a user study using Amazon MTurk~\cite{mturk} on the Vid4~\cite{vid4} and REDS4~\cite{Nah_2019_CVPR_Workshops_REDS} datasets.
We evaluate the COMISR model against all other methods using videos compressed with CRF25.
In each experiment, two videos generated by the COMISR model and other methods are presented side by side and 
each user is asked ``which video looks better?''
For the Vid4 and REDS4 datasets, all the test videos are used for the user study.
For each of the video pairs, we assign to 20 different raters.
The aggregated results are shown in Figure~\ref{fig:userstudy}.
